\documentclass[10pt]{article} 
\usepackage[preprint]{tmlr}

\usepackage{hyperref}
\usepackage{amsmath}
\usepackage{url}
\usepackage{natbib}
\usepackage{graphicx}
\usepackage{graphics}
\usepackage{multirow}
\usepackage{bbding}
\usepackage{amsfonts}
\usepackage{subcaption}
\usepackage{anyfontsize}
\newsavebox\CBox

\title{SupEuclid: Extremely Simple, High Quality OoD Detection with Supervised Contrastive Learning and Euclidean Distance}

\author{\name Jarrod Haas \email jhaas@sfu.ca \\
      \addr SARlab, Department of Engineering Science\\
      Simon Fraser University; Digitalist Group Canada
      }

\def\month{12}  
\def\year{2022} 
\def\openreview{\url{https://openreview.net/forum?id=fjkN5Ur2d6}} 

\begin{document}
\maketitle

\begin{abstract}
Out-of-Distribution (OoD) detection has developed substantially in the past few years, with available methods approaching, and in a few cases achieving, perfect data separation on standard benchmarks. These results generally involve large or complex models, pretraining, exposure to OoD examples or extra hyperparameter tuning. Remarkably, it is possible to achieve results that can exceed many of these state-of-the-art methods with a very simple method. We demonstrate that ResNet18 trained with Supervised Contrastive Learning (SCL) produces state-of-the-art results out-of-the-box on near and far OoD detection benchmarks using only Euclidean distance as a scoring rule. This may obviate the need in some cases for more sophisticated methods or larger models, and at the very least provides a very strong, easy to use baseline for further experimentation and analysis.

\end{abstract}

\section{Introduction}

Neural networks often give unreliable confidence scores when receiving images outside the training distribution, hindering their application in real-world settings where reliable failure modes are critical \citep{FailingLoudly:AnEmpiricalStudyofMethodsforDetectingDatasetShift, RobustOut-of-distributionDetectioninNeuralNetworks, henne2020benchmarking}. Much work has been done to address this shortcoming. In a few short years, early and popular Monte-Carlo Dropout (MCD) and model ensemble methods have become obsolete \citep{gal2016dropout, lakshminarayanan2017simple}. These have been supplanted by a diverse array of methods that can roughly be characterized as modifying and measuring feature space. Results across OoD benchmarks have improved substantially via techniques that include large or complex models, outlier exposure, pretraining, sophisticated loss functions or scoring functions and extra hyperparameters. We demonstrate that, remarkably, it is possible to surpass many of these methods and approach near-perfect benchmark scores with only a ResNet18 trained out-of-the-box with Supervised Contrastive Loss (SCL) \citep{khosla2021supervised} and Euclidean distance as a scoring rule. 

Our contribution is as follows:

\begin{itemize}
\item \kern-4pt State-of-the-art OoD detection using only ResNet18 trained with out-of-the-box Supervised Contrastive Loss (code available), and requiring no outlier exposure, complicated hyperparameter tuning, pretraining, additional test-time computation, ensembling or complex scoring rules. 

\end{itemize}

\section{Methodology}
\label{gen_inst}

A standard classification model maps images to classes $f:x\rightarrow y,$  where $x \in X$ is the set of images, and $y\in \left \{  1,...,k\right \}$ is the set of $k$ possible classes. The model is composed of a feature extractor which embeds images into feature space embeddings $z$ in $\mathbb{R}^d$ where $d$ is the size of the feature vector, and a linear decision classifier that transforms $z$ into a vector of length $k$, known as the logits. These logits are input to a softmax function to produce a probability over classes over which the $argmax$ function identifies the class prediction $y$. We note that the set of images can be broken into $X \in X_{in}$ which are images drawn from the same distribution as the training data, and $X \in X_{out}$, which are all other images.

Instead of the standard Cross Entropy (CE) loss, we use Supervised Contrastive Loss. To evaluate models, we merge ID and OoD images into a single test set. OoD performance is then a binary classification task, where we measure how well OoD images can be separated from ID images using a score derived from our model. Our score in this case is simply the minimum Euclidean distance of each image's feature vector, $z$, to all class means. Task performance is scored via AUROC and FPR95.

\begin{table}
\fontsize{7pt}{7pt}\selectfont
\resizebox{\columnwidth}{!}{
\begin{tabular}{lccccc}
\hline \\
\textbf{Method}                      & \textbf{Network} & \multicolumn{2}{c}{\textbf{SVHN}} & \multicolumn{2}{c}{\textbf{CIFAR100}} \\\\ \hline \\
Mahalanobis \citeyear{lee2018simple}                          & ResNet34         & 99.1            & 9.24            & 90.5              & N/A               \\
LogitNorm \citeyear{wei2022mitigating}                      & ResNet18         & 97.1            & N/A             & 91.0              & N/A               \\
L2Mag60 \citeyear{haas2023simple}                             & ResNet18         & 97.4            & N/A             & 88.7              & N/A               \\
FeatureNorm \citeyear{yu2023block}                         & ResNet18         & 98.7            & 7.13            & 74.5              & N/A               \\
KNN+ \citeyear{sun2022outofdistribution}                                & ResNet18         & 99.5            & 2.42            & N/A               & 38.8              \\
SSD+ \citeyear{sehwag2021ssd}                                & ResNet50         & 99.9            & 2.00            & 93.4              & 38.5              \\
CSI (Unsupervised) \citeyear{tack2020csi}                   & ResNet18         & 99.8            & N/A             & 89.2              & N/A               \\
Gram \citeyear{sastry2020detecting}                                 & ResNet34         & 99.5            & N/A             & 79.0              & N/A               \\
Contrastive (SimCLR + Mahalanobis) \citeyear{winkens2020contrastive}   & WideResNet50\_4  & 99.5            & 2.8             & 92.9              & 39.9              \\
ERD++ \citeyear{tack2020csi}                                & 3x ResNet20      & 1.00            & N/A             & 95.0              & N/A               \\ \\  
Pretrained Transformer + Mahalanobis & R50+ViT-B\_16 \citeyear{fort2021exploring}    & 99.9            & 1.9             & 98.5              & 6.89              \\  \hline \\
SupEuclid (Ours)                     & ResNet18         & 99.6            & 2.2             & 92.1              & 31.5              \\  \hline
\end{tabular}}

\caption{AUROC / FPR95 Results for SupEuclid compared to an array of popular, strong baselines. Baselines are selected to be in the same model class (i.e. ResNet18) wherever possible. The transformer scores are reported as an example of possible results, regardless of model size. All metrics are reported for models trained on CIFAR10, and tested on SVHN and CIFAR100 as OoD datasets. Our scores are the mean of fifteen ResNet18 models, each randomly initialized prior to training. Our method matches or exceeds others in it's class, particularly when factoring in FPR95 scores on CIFAR100, as is the simplest to implement and use. More comprehensive scores are reported in our forthcoming paper.}
\label{tab:main_results}
\end{table}

\section{Experiments and Conclusion}

All metrics are calculated from fifteen standard ResNet18 models initialized randomly and trained with SCL loss on CIFAR10. The training code can be found online \footnote{https://github.com/HobbitLong/SupContrast/tree/master}, which is the same repo noted on the original Supervised Contrastive Learning paper. We train for 500 epochs instead of 1000, but use the exact default training and data augmentation parameters (as noted in the github repo readme and within the code): an SGD optimizer with an initial learning rate of 0.5, gamma=0.1, and with the default cosine annealing parameters. 

Our results in Table \ref{tab:main_results} demonstrate that an out-of-the-box ResNet18 trained with SCL and using Euclidean distance to class means as a scoring rule produces results that are competitive with or exceed many state-of-the-art baselines.

\newpage
\bibliography{main}

\begin{thebibliography}{16}
\providecommand{\natexlab}[1]{#1}
\providecommand{\url}[1]{\texttt{#1}}
\expandafter\ifx\csname urlstyle\endcsname\relax
  \providecommand{\doi}[1]{doi: #1}\else
  \providecommand{\doi}{doi: \begingroup \urlstyle{rm}\Url}\fi

\bibitem[Chen et~al.(2020)Chen, Li, Wu, Liang, and
  Jha]{RobustOut-of-distributionDetectioninNeuralNetworks}
Jiefeng Chen, Yixuan Li, Xi~Wu, Yingyu Liang, and Somesh Jha.
\newblock Robust out-of-distribution detection in neural networks.
\newblock \emph{CoRR}, abs/2003.09711, 2020.
\newblock URL \url{https://arxiv.org/abs/2003.09711}.

\bibitem[Fort et~al.(2021)Fort, Ren, and Lakshminarayanan]{fort2021exploring}
Stanislav Fort, Jie Ren, and Balaji Lakshminarayanan.
\newblock Exploring the limits of out-of-distribution detection, 2021.

\bibitem[Gal and Ghahramani(2016)]{gal2016dropout}
Yarin Gal and Zoubin Ghahramani.
\newblock Dropout as a bayesian approximation: Representing model uncertainty
  in deep learning.
\newblock In \emph{international conference on machine learning}, pages
  1050--1059. PMLR, 2016.

\bibitem[Haas et~al.(2023)Haas, Yolland, and Rabus]{haas2023simple}
Jarrod Haas, William Yolland, and Bernhard Rabus.
\newblock Simple high quality ood detection with l2 normalization, 2023.

\bibitem[Henne et~al.(2020)Henne, Schwaiger, Roscher, and
  Weiss]{henne2020benchmarking}
Maximilian Henne, Adrian Schwaiger, Karsten Roscher, and Gereon Weiss.
\newblock Benchmarking uncertainty estimation methods for deep learning with
  safety-related metrics.
\newblock In \emph{SafeAI@ AAAI}, pages 83--90, 2020.

\bibitem[Khosla et~al.(2021)Khosla, Teterwak, Wang, Sarna, Tian, Isola,
  Maschinot, Liu, and Krishnan]{khosla2021supervised}
Prannay Khosla, Piotr Teterwak, Chen Wang, Aaron Sarna, Yonglong Tian, Phillip
  Isola, Aaron Maschinot, Ce~Liu, and Dilip Krishnan.
\newblock Supervised contrastive learning, 2021.

\bibitem[Lakshminarayanan et~al.(2017)Lakshminarayanan, Pritzel, and
  Blundell]{lakshminarayanan2017simple}
Balaji Lakshminarayanan, Alexander Pritzel, and Charles Blundell.
\newblock Simple and scalable predictive uncertainty estimation using deep
  ensembles.
\newblock \emph{Advances in neural information processing systems}, 30, 2017.

\bibitem[Lee et~al.(2018)Lee, Lee, Lee, and Shin]{lee2018simple}
Kimin Lee, Kibok Lee, Honglak Lee, and Jinwoo Shin.
\newblock A simple unified framework for detecting out-of-distribution samples
  and adversarial attacks, 2018.

\bibitem[Rabanser et~al.(2018)Rabanser, Günnemann, and
  Lipton]{FailingLoudly:AnEmpiricalStudyofMethodsforDetectingDatasetShift}
Stephan Rabanser, Stephan Günnemann, and Zachary~C. Lipton.
\newblock Failing loudly: An empirical study of methods for detecting dataset
  shift, 2018.
\newblock URL \url{https://arxiv.org/abs/1810.11953}.

\bibitem[Sastry and Oore(2020)]{sastry2020detecting}
Chandramouli~Shama Sastry and Sageev Oore.
\newblock Detecting out-of-distribution examples with in-distribution examples
  and gram matrices, 2020.

\bibitem[Sehwag et~al.(2021)Sehwag, Chiang, and Mittal]{sehwag2021ssd}
Vikash Sehwag, Mung Chiang, and Prateek Mittal.
\newblock Ssd: A unified framework for self-supervised outlier detection, 2021.

\bibitem[Sun et~al.(2022)Sun, Ming, Zhu, and Li]{sun2022outofdistribution}
Yiyou Sun, Yifei Ming, Xiaojin Zhu, and Yixuan Li.
\newblock Out-of-distribution detection with deep nearest neighbors, 2022.

\bibitem[Tack et~al.(2020)Tack, Mo, Jeong, and Shin]{tack2020csi}
Jihoon Tack, Sangwoo Mo, Jongheon Jeong, and Jinwoo Shin.
\newblock Csi: Novelty detection via contrastive learning on distributionally
  shifted instances, 2020.

\bibitem[Wei et~al.(2022)Wei, Xie, Cheng, Feng, An, and Li]{wei2022mitigating}
Hongxin Wei, Renchunzi Xie, Hao Cheng, Lei Feng, Bo~An, and Yixuan Li.
\newblock Mitigating neural network overconfidence with logit normalization.
\newblock \emph{arXiv preprint arXiv:2205.09310}, 2022.

\bibitem[Winkens et~al.(2020)Winkens, Bunel, Roy, Stanforth, Natarajan, Ledsam,
  MacWilliams, Kohli, Karthikesalingam, Kohl, Cemgil, Eslami, and
  Ronneberger]{winkens2020contrastive}
Jim Winkens, Rudy Bunel, Abhijit~Guha Roy, Robert Stanforth, Vivek Natarajan,
  Joseph~R. Ledsam, Patricia MacWilliams, Pushmeet Kohli, Alan
  Karthikesalingam, Simon Kohl, Taylan Cemgil, S.~M.~Ali Eslami, and Olaf
  Ronneberger.
\newblock Contrastive training for improved out-of-distribution detection,
  2020.

\bibitem[Yu et~al.(2023)Yu, Shin, Lee, Jun, and Lee]{yu2023block}
Yeonguk Yu, Sungho Shin, Seongju Lee, Changhyun Jun, and Kyoobin Lee.
\newblock Block selection method for using feature norm in out-of-distribution
  detection, 2023.

\end{thebibliography}

\end{document}